\documentclass[10pt,twocolumn,letterpaper]{article}

\usepackage{cvpr}              

\usepackage{multirow}
\usepackage{graphicx}
\usepackage{amsmath}
\usepackage{amssymb}
\usepackage{booktabs}
\usepackage{array}
\usepackage{amsfonts}
\usepackage{bbm}
\usepackage{colortbl}

\usepackage[accsupp]{axessibility}  

\usepackage[pagebackref,breaklinks,colorlinks]{hyperref}

\usepackage[capitalize]{cleveref}
\crefname{section}{Sec.}{Secs.}
\Crefname{section}{Section}{Sections}
\Crefname{table}{Table}{Tables}
\crefname{table}{Tab.}{Tabs.}

\def\cvprPaperID{1311} 
\def\confName{CVPR}
\def\confYear{2022}

\begin{document}

\title{Learning Second Order Local Anomaly for General Face Forgery Detection}

\author{Jianwei Fei$^{1}$\footnotemark[1]~~ Yunshu Dai$^{1}$\footnotemark[1]~~ Peipeng Yu$^2$~~ Tianrun Shen$^3$~~ Zhihua Xia$^{2,1}$\footnotemark[2]~~ Jian Weng$^2$\\
$^1$Nanjing University of Information Science and Technology~ 
$^2$Jinan University~ 
$^3$Nanjing University\\

{\tt\small {\{fjw826244895,daiyunshu0102,ypp865,xia\_zhihua\}@163.com}~~ }\\ 
{\tt\small tiruns@yeah.net~~ cryptjweng@gmail.com}
}

\maketitle
\renewcommand{\thefootnote}{\fnsymbol{footnote}} 
\footnotetext[1]{These authors contributed equally to this work.} 
\footnotetext[2]{Corresponding author.} 
\begin{abstract}
In this work, we propose a novel method to improve the generalization ability of CNN-based face forgery detectors. Our method considers the feature anomalies of forged faces caused by the prevalent blending operations in face forgery algorithms. Specifically, we propose a weakly supervised Second Order Local Anomaly (SOLA) learning module to mine anomalies in local regions using deep feature maps. SOLA first decomposes the neighborhood of local features by different directions and distances and then calculates the first and second order local anomaly maps which provide more general forgery traces for the classifier. We also propose a Local Enhancement Module (LEM) to improve the discrimination between local features of real and forged regions, so as to ensure accuracy in calculating anomalies. Besides, an improved Adaptive Spatial Rich Model (ASRM) is introduced to help mine subtle noise features via learnable high pass filters. With neither pixel level annotations nor external synthetic data, our method using a simple ResNet18 backbone achieves competitive performances compared with state-of-the-art works when evaluated on unseen forgeries.
\end{abstract}

\section{Introduction}
\label{sec:Intro}
Recent progress in face synthesis technologies allows the low-entry production of sophisticated fake facial content which causes severe trust issues. Concerns are growing over the nefarious use of such face forgery technologies. To deal with this problem, 
\begin{figure}[htbp]
  \centering
   \includegraphics[width=\linewidth]{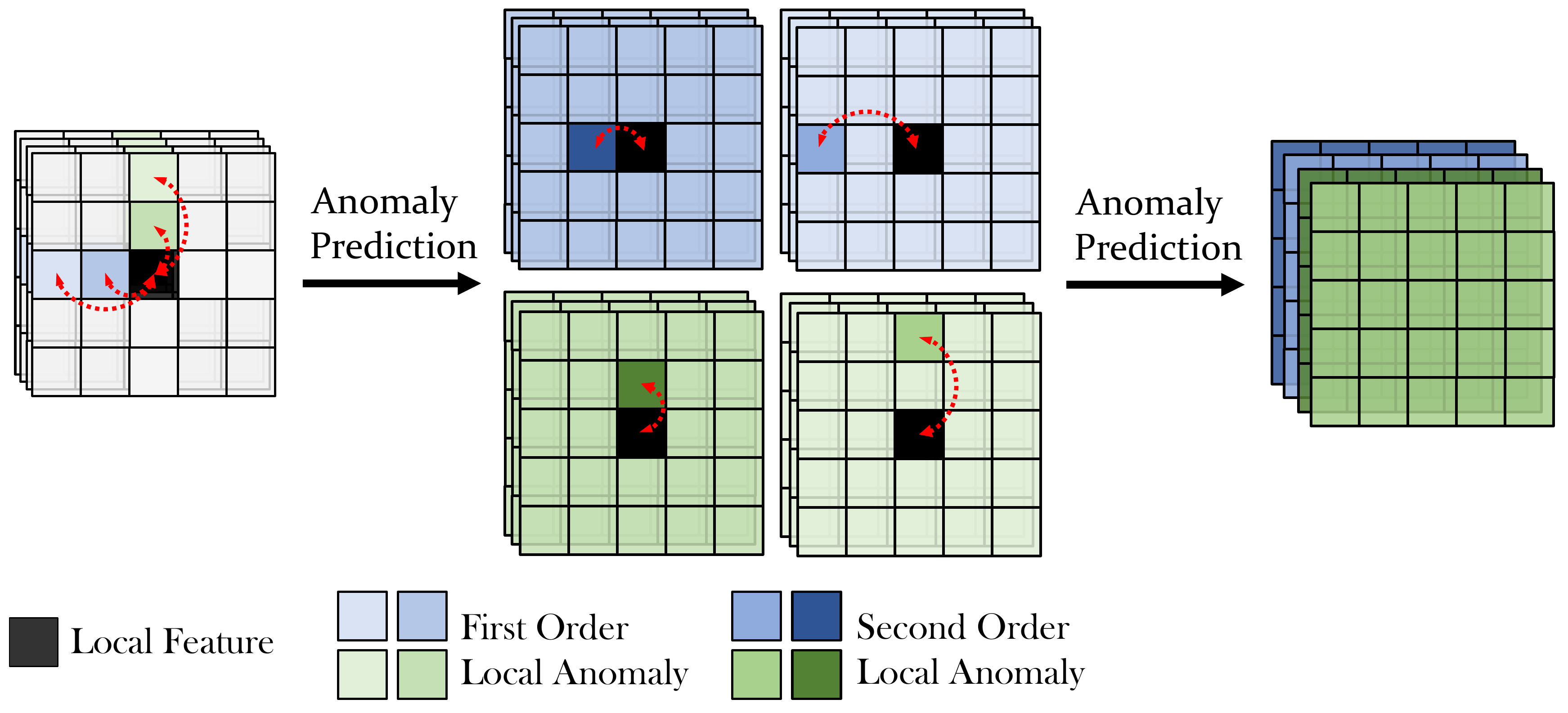}
   \caption{Second Order Local Anomaly (SOLA) learning module decomposes the neighborhood and predicts the first and second order anomaly maps.}
   \label{fig:sola}
\end{figure}
many methods have been developed to detect face forgeries using different traces, such as obvious visual artifacts \cite{li2019exposing,schwarcz2021finding,matern2019exploiting}, frequency domain cues \cite{durall2020watch,frank2020leveraging,qian2020thinking,li2021frequency}, temporal anomalies \cite{sabir2019recurrent,masi2020two,wu2020sstnet,chen2020fsspotter,ru2021bita}, or multimodality conflicts \cite{chugh2020not,mittal2020emotions,agarwal2020detecting}. But these traces are not so generalizable that the detectors may fail when encountering unseen forgeries. Thus recent works are looking for more universal forgery traces, so as to boost the generalizability.

All the face forgery algorithms, whether face swapping or reenactment, need to blend the forged regions into the original background. The two parts inevitably hold different features, especially in high frequency regions, resulting in anomaly in forged images.
On such observation, some methods are proposed to capture the feature inconsistency \cite{li2020face,chen2021local,zhao2021learning}. \cite{zhao2021learning} propose a patch wise consistency learning method to mine anomalies in forged face images and succeed in generalizing to unseen forgeries.
However, these works treat regions with various distances equally, which violates the fact that natural images hold different dependencies of short and long distances.
Moreover, they usually require either pixel level annotations or external synthetic data. Although pixel level annotations can be created in lab environments, the need for them limits the usage of forged faces transmitted in the real world.

In this work, we focus on capturing forgery traces from the perspective of local anomaly. 
Specifically, given a face image, we extract its deep features with a CNN backbone, and divide the neighborhood of a local feature into 4 groups by different orientations and distances, thereby modeling more fine-grained local anomalies and generating 4 groups of anomaly maps. 
These anomaly maps are then decomposed again to calculate the second order anomaly which have a wide range of response to the first order anomaly. The overview is shown in Fig. \ref{fig:sola}, and more details are in Sec. \ref{sec:sola}.
Our method requires neither pixel level annotations nor external data but achieves great cross-domain performance with a small backbone.


\begin{figure*}[htbp]
  \centering
   \includegraphics[width=\linewidth]{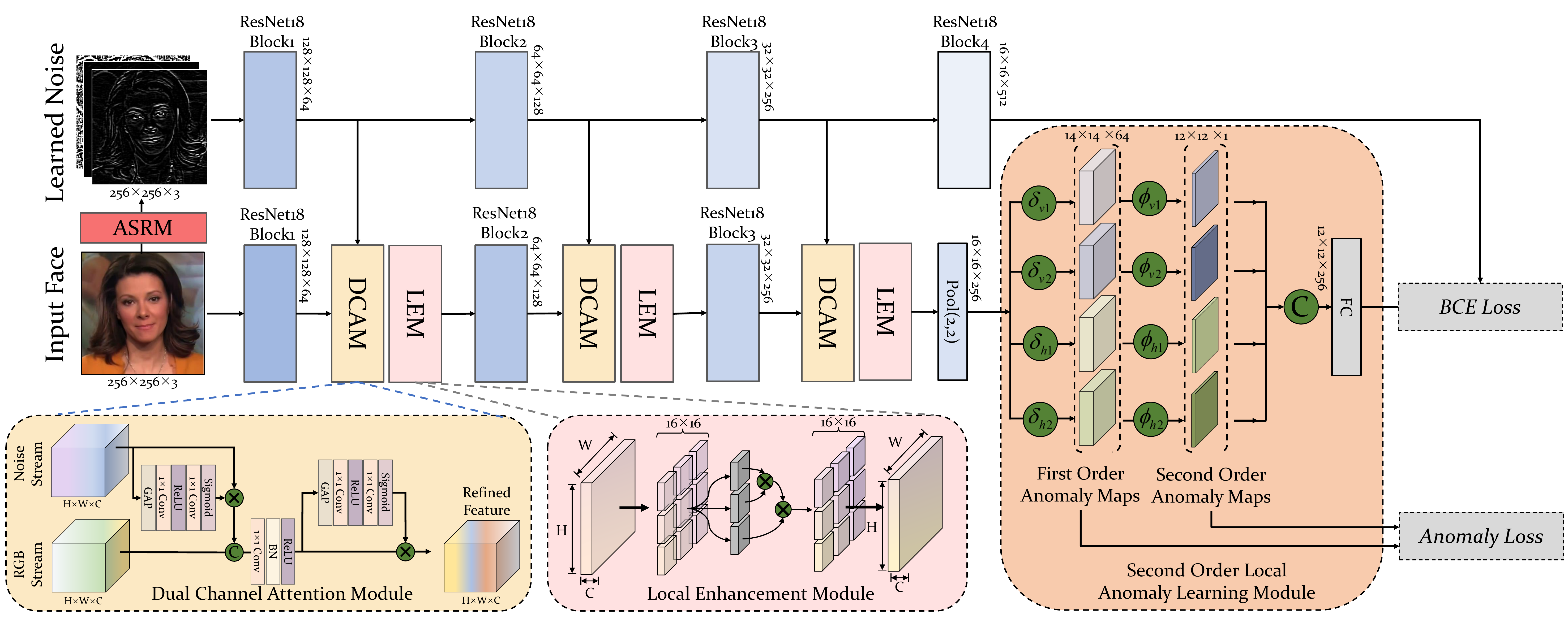}
   \caption{Overview of the proposed method (with ResNet18 backbone).}
   \label{fig:overview}
\end{figure*}

Our contributions can be summarized into four-fold: 
\vspace{-2mm}
\begin{itemize}
  \setlength{\itemsep}{0pt}
  \setlength{\parsep}{0pt}
  \setlength{\parskip}{0pt}
  \item We propose a second order local anomaly (SOLA) learning module for isolating forgery faces. 
  SOLA calculates anomalies of a local region in fine-grained fashion and magnifies the forgery traces via calculating the second order anomalies.
  Classifiers using the anomaly features are proved to have remarkable generalization on unseen forgeries.
  \item We propose a local enhancement module (LEM) that can be inserted after different stages of the backbone to ensure the discrimination between real and forged regions in deep local features.
  \item We design an adaptive spatial rich model (ASRM) which inherits the high pass peculiarity of SRM filters in forensics and is able to adaptively adjust the filters according to data. 
  \item Our method requires neither pixel level annotations nor external synthetic data to improve the generalization of face forgery detectors.
\end{itemize}


\section{Related work}
\label{sec:related work}
Face forgeries can be seen as a game of AI versus AI since most detection technologies are deep learning based. Although these methods perform well under in-domain evaluations, they often suffer performance degeneration on unseen forgeries. 

\textbf{Generalization to Unseen Forgeries.}
Many efforts are made to improve the generalizability of detectors such as combining auxiliary localization tasks to guide the network focus more on forged regions \cite{li2020face,dang2020detection}, improving cross entropy loss with metric learning for better class discrimination \cite{li2021frequency,masi2020two}, and introducing domain adaptation to alleviate overfitting on a single domain \cite{rao2021self, sun2021domain}.
Recently, methods using frequency cues \cite{qian2020thinking,liu2021spatial} or combining features from different domains also perform well \cite{luo2021generalizing,masi2020two} in cross-domain evaluations. 
In \cite{li2021frequency}, one branch process the RGB input, while another uses the DCT transformation to extract high frequency features from different bands. Outputs of the two branches are fused to form more generalized forgery features. 
Many works similarly fuse features from RGB and frequency domains for more general representation \cite{han2021fighting,luo2021generalizing,chen2021local,jia2021inconsistency}. However, their methods of extracting high frequency features can not fit the data adaptively to capture the most discriminative features.


\textbf{Anomaly-based Face Forgery Detection.}
Some works introduce the idea of anomaly detection into image forensics \cite{cozzolino2016single,fatemifar2021client} as well as face forgery detection \cite{ijcai2021-102,khalid2020oc}, and achieve good generalization. For example, Wu et al. \cite{wu2019mantra} calculate the differences between local feature and dominant feature of an image, turning the forgery localization into local anomaly detection. Hu et al. \cite{ijcai2021-102} propose a dynamic inconsistency aware network to capture both global inconsistency of adjacent frames and local inconsistency of key regions within frames.

Recent methods similarly utilize the patch level anomalies for face forgery detection. 
They are mainly based on the observation that face forgery algorithms always blend forged regions with original face context, resulting in pixel statistical property anomalies within one image \cite{zhao2021learning}.
An example of the pixel statistical property is camera noise \cite{cozzolino2019noiseprint} which is a high frequency feature left during the imaging process and has been used for detecting image integrity \cite{verdoliva2019extracting}.
\cite{zhao2021learning} and \cite{chen2021local} propose to divide deep feature maps into patches, and calculate the similarities between patches to form a more generalized pattern that indicates the image integrity. 
They achieve both great performances under in-domain and cross-domain evaluations. 
Unlike these methods that explore global level consistency, our method explores more fine-grained local anomaly instead.



\section{Method}
\label{sec:Method}
The overview of our method is illustrated in Fig. \ref{fig:overview}. 
Given an input face image, it is fed to the RGB branch, as well as a parallel noise branch whose first layer is our Adaptive Spatial Rich Model (ASRM). 
These intermediate features extracted from these two branches are fused through Dual Channel Attention Module (DCAM) after each block.
Then the Local Enhancement Module (LEM) divide the fused feature maps into patches and enhance the category attributes of each local feature inside. 
Finally, the Second Order Local Anomaly (SOLA) learning module calculates fine-grained local anomalies of the first and second order, and predicts whether the input is forged.

  
\subsection{Adaptive SRM in Noise Branch}
\label{sec:asrm}
\textbf{Motivation.}
Learning from the RGB data is insufficient to capture high frequency features that are crucial for image forensics. 
To solve this problem, SRM (Spatial Rich Model) \cite{fridrich2012rich} has been widely applied for preprocessing \cite{han2021fighting} and layer initialization \cite{rao2016deep} to extract high frequency noise.
But if used for preprocessing, the handcrafted filters in SRM can not adaptively update to fit the data. Besides, if used for initialization, the backpropagation will adjust the filters and break their high pass peculiarity. 
In this work, we address this dilemma by introducing a constraint to let SRM adaptively update while maintaining their high frequency learning ability.
\begin{figure}[hbp]
  \centering
   \includegraphics[width=0.7\linewidth]{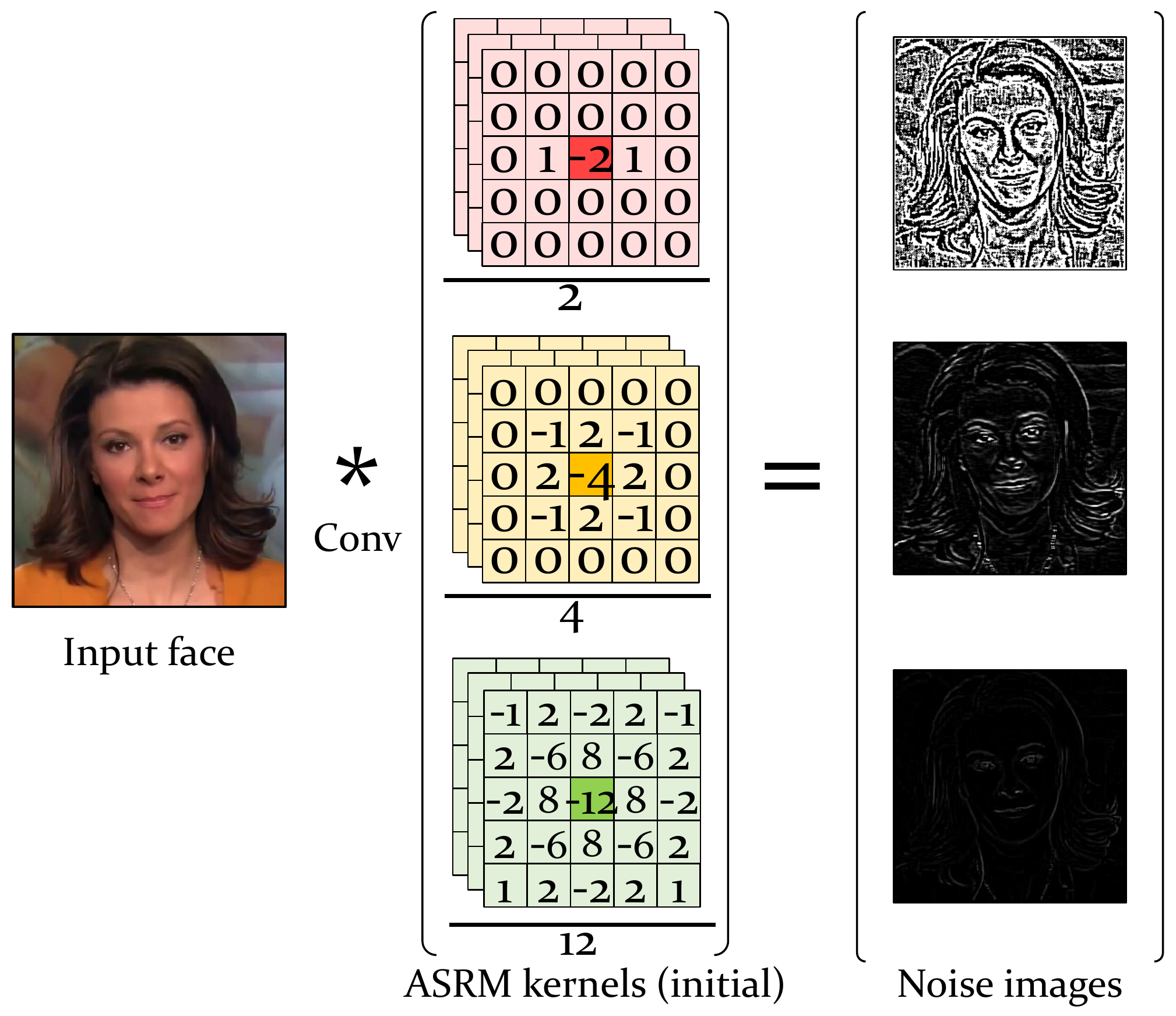}
   \caption{The Adaptive Spatial Rich Model (ASRM) convolution.}
   \label{fig:srm}
\end{figure}

\textbf{Design of ASRM.}
We start with original SRM following \cite{fridrich2012rich}. SRM models the noise residuals $R_{i,j}$ using high pass filters by:
\begin{equation}
  \setlength{\abovedisplayskip}{4pt}
  R_{i,j} = P(\mathcal{N}_{i,j})-I_{i,j},
  \setlength{\belowdisplayskip}{4pt}
  \label{eq:srm}
\end{equation}
where $\mathcal{N}_{i,j}$ is the neighborhood of center pixel $I_{i,j}$, $P(\cdot)$ is a predictor of $I_{i,j}$ based on $\mathcal{N}_{i,j}$. The residual is then quantitated by $q$, following with rounding and truncation: 
\begin{equation}
  \setlength{\abovedisplayskip}{2pt}
  R_{i,j} \gets trun(round(\frac{R_{i,j}}{q})).
  \setlength{\belowdisplayskip}{2pt}
  \label{eq:quan}
\end{equation}

We can see that the $P(\cdot)$ can be realized by a standard convolution. Following \cite{bayar2018constrained}, let $\emph{\textbf{I}} * \emph{\textbf{w}}$ denotes a standard convolution on input image $\emph{\textbf{I}}$ with kernel $\emph{\textbf{w}}$, and $o$ notes an impulse filter with central value being 1 and others being 0, we then have:
\begin{equation}
  \setlength{\abovedisplayskip}{3pt}
  \emph{\textbf{h}} = \emph{\textbf{I}} * \emph{\textbf{w}} = \emph{\textbf{I}} * \hat{\emph{\textbf{w}}} - \emph{\textbf{I}} = \emph{\textbf{I}} * (\hat{\emph{\textbf{w}}} - o ),
  \setlength{\belowdisplayskip}{3pt}
  \label{eq:bayer}
\end{equation}
where the central value of $\hat{\emph{\textbf{w}}}$ is 0. Then the central value of $\emph{\textbf{w}}$ is -1 and $\emph{\textbf{I}} * \emph{\textbf{w}}$ is equal to Eq. \ref{eq:srm}.

We select 3 out of 30 filters in SRM as they have been experimentally proved to be effective enough \cite{zhou2018learning}. 
The filters are repeated 3 times for RGB images as shown in Fig.\ref{fig:srm}. 
To let their center values be -1, we quantitate them by 2, 4, and 12 respectively. 
We can see that after quantitation, the sum of the remaining elements is 1 which perfectly fit the requirements of Eq. \ref{eq:bayer}.
To allow the filters learnable during network training while keep their peculiarity in high pass filtering, we reset their central element to -1 and normalize the remaining elements to force their sum to be 1:
\begin{equation}
  \begin{cases}
    \setlength{\abovedisplayskip}{5pt}
    \emph{\textbf{w}}_{0,0} = -1,\\
    \sum\nolimits_{i\neq 0,j\neq 0} \emph{\textbf{w}}_{i,j} = 1,
    \setlength{\belowdisplayskip}{5pt}
  \end{cases}
  \label{eq:constraint}
\end{equation}
where $i, j$ denotes the index of elements in kernels and $(0,0)$ is the central one. We execute this constraint after each backpropagation.

ASRM is the first layer used for the preprocessing of the noise branch. Different from using fixed SRM kernels, ASRM enables the kernels to adaptively update during the backpropagation and extract discriminative noise features that can not be directly learned from RGB data.

\begin{figure*}[htp]
  \centering
   \includegraphics[width=0.95\linewidth]{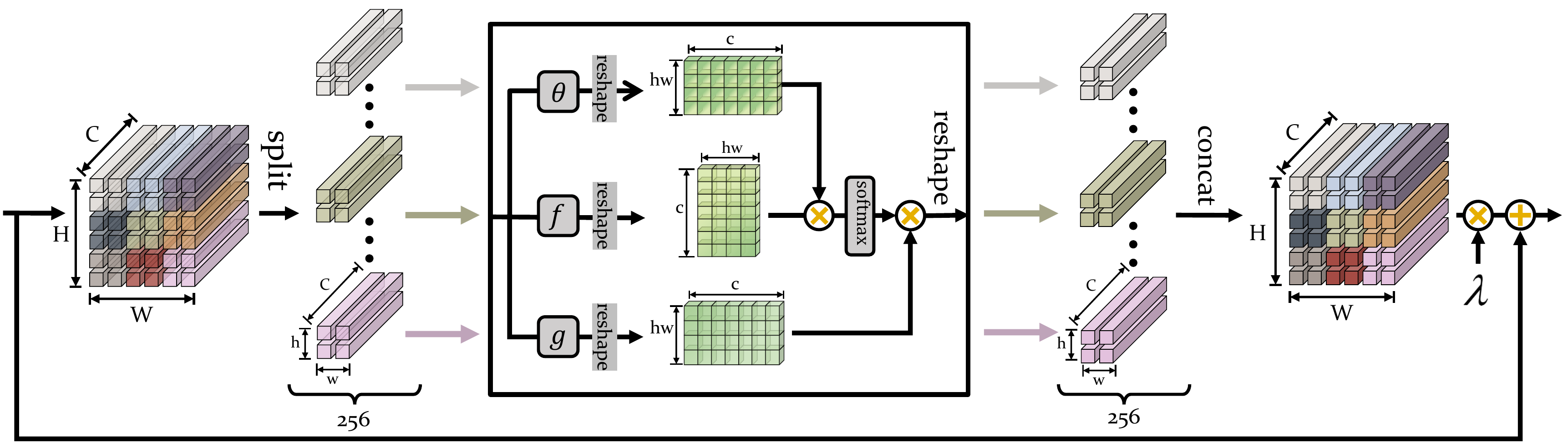}
   \caption{The Local Enhancement Module (LEM).}
   \label{fig:lem}
\end{figure*}
The noise features extracted by ASRM provide subtle high frequency cues, but different channels of the feature maps have different contributions to generalization. Besides, different channels in the fused features are of different importance in forgery representations. Thus, we propose the Dual Channel Attention Module (DCAM) to fuse the intermediate features of RGB and noise branches. DCAM uses channel attention \cite{woo2018cbam} twice to emphasize channels of more importance before and after the fusion. 

\subsection{Local Enhancement Module}
  \label{sec:lem}
  SOLA calculates local anomalies on such an assumption: each pixel in the feature maps is a local feature vector associated with a corresponding local region whose size is the ratio of input size to feature size. 
  However, a feature vector is associated with a larger region due to the expansion of receptive field after several layers. 
  The feature vectors are thereby affected by information of outer regions and lose their discrimination in representing local regions.
  We notice that some works have been proposed to capture long range dependencies in images \cite{wang2018non, zhang2019self} with self attention. 
  Following this idea, we propose the Local Enhancement Module (LEM) to improve the discrimination of local feature vectors by assigning them with different weights.

  Specifically, LEM takes as input the intermediate feature maps $\textbf{F}^l \in \mathbb{R}^{h_l \times w_l \times c_l}$ where $l$ denotes the hierarchy. We spatially divide $\textbf{F}^l$ into 16 $\times$ 16 non-overlapping patches $\textbf{p}^{l}_{k},k=1,2,...,256$ since size of the final feature maps is $16 \times 16$.
  $\textbf{p}^{l}_{k}$ is transformed using embedding functions $\theta $ and $f$ to calculate a weight matrix $\textbf{W}^{l}_{k}$ by:
  \begin{equation}
    \small
    \setlength{\abovedisplayskip}{3pt}
    \textbf{W}^{l}_{k} = softmax (\theta  (\textbf{p}^{l}_{k})^T \otimes  f(\textbf{p}^{l}_{k})).
    \label{eq:lem}
    \setlength{\belowdisplayskip}{3pt}
  \end{equation}
  The elements in $\textbf{W}^{l}_{k}$ indicate the relevance between local features in $\textbf{p}^{l}_{k}$ and separate local features by assigning different weights, given that local features of forged regions are more relevant to that of forged regions and vice versa. 
  $\textbf{p}^{l}_{k}$ is also transformed by $g$ and then enhanced by:
  \begin{equation}
    \small
    \setlength{\abovedisplayskip}{3pt}
    \hat{\textbf{p}}^{l}_{k} = \textbf{W}^{l}_{k} \otimes g (\textbf{p}^{l}_{k}).
    \label{eq:lem}
    \setlength{\belowdisplayskip}{3pt}
  \end{equation}
  As illustrated in Fig.\ref{fig:lem}, LEM concats $\hat{\textbf{p}}^{l}_{k}$ to form the enhanced feature maps $\hat{\textbf{F}}^l$ which is then multiplied by a learnable scale parameter $\lambda$ and added with the input feature.

  \subsection{Second Order Local Anomaly Learning}
  \label{sec:sola}

  In this work, we propose the Second Order Local Anomaly (SOLA) learning module to detect face forgeries by fine-grained local anomalies. 
  Let $\textbf{F}_{i,j}\in \mathbb{R}^{C}$ denotes the $(i^{th},j^{th})$ feature vector that corresponds to a local region in the input image.
  To obtain more fine-grained representation of local anomalies, SOLA first decomposes the neighborhood of $\textbf{F}_{i,j}$ into horizontal/vertical and nearest/next nearest neighbors by direction and distance respectively.  
  In practice, we only predict the anomalies of upper and left neighbors due to the symmetry. Then the first order anomaly maps of each $\textbf{F}_{i,j}$ are predicted by Eq. \ref{eq:first order}:
  \begin{equation}
  \begin{cases}
    \setlength{\abovedisplayskip}{0pt}
    \mathcal{M}_{v1,(i,j)} = \delta_{v1}(\textbf{F}_{i,j}, \textbf{F}_{i,j-1}),\\
    \mathcal{M}_{v2,(i,j)} = \delta_{v2}(\textbf{F}_{i,j}, \textbf{F}_{i,j-2}),\\
    \mathcal{M}_{h1,(i,j)} = \delta_{h1}(\textbf{F}_{i,j}, \textbf{F}_{i-1,j}),\\
    \mathcal{M}_{h2,(i,j)} = \delta_{h2}(\textbf{F}_{i,j}, \textbf{F}_{i-2,j}),
    \setlength{\belowdisplayskip}{0pt}
  \end{cases}
  \label{eq:first order}
  \end{equation}
  where $\delta(\cdot)$ is the anomaly predictor realized by $1 \times 2$ or $2 \times 1$ convolutions (dilation = 2 for next nearest neighbors, otherwise 1), $v$ and $h$ denote vertical and horizontal neighbors, 1 and 2 denote the nearest and next nearest neighbors. 
  One kernel in $\delta(\cdot)$ can predict only one anomaly map, thus we set each $\delta(\cdot)$ to contain 64 kernels so that it can generate 64 anomaly maps to capture multiple anomalies. 

  Inspired by the idea of the Laplace operator that computes the second derivative, we propose the second order local anomaly learning to obtain magnified responses to the first order anomalies. To this end, SOLA applies the local decomposition again on the first order anomaly maps and calculates the second order local anomaly maps by Eq. \ref{eq:second order}. 
  A pixel in the second order anomaly maps is responded to a region in the first order anomaly maps so as to capture a wider range of fine-grained anomalies.
  \begin{equation}
    \begin{cases}
      \setlength{\abovedisplayskip}{0pt}
      \mathcal{M}^{\prime}_{v1,(i,j)} = \phi_{v1}(\mathcal{M}_{v1,(i,j)}, \mathcal{M}_{v1,(i,j-1)}),\\
      \mathcal{M}^{\prime}_{v2,(i,j)} = \phi_{v2}(\mathcal{M}_{v2,(i,j)}, \mathcal{M}_{v2,(i,j-2)}),\\
      \mathcal{M}^{\prime}_{h1,(i,j)} = \phi_{h1}(\mathcal{M}_{h1,(i,j)}, \mathcal{M}_{h1,(i-1,j)}),\\
      \mathcal{M}^{\prime}_{h2,(i,j)} = \phi_{h2}(\mathcal{M}_{h2,(i,j)}, \mathcal{M}_{h2,(i-2,j)}),
      \setlength{\belowdisplayskip}{0pt}
    \end{cases}
    \label{eq:second order}
    \end{equation}
  $\phi(\cdot)$ denotes the second order anomaly predictor that is similar to $\delta(\cdot)$ but contains only one single kernel. Finally, we stack 4 outputs of $\phi(\cdot)$ to form a 4 channel anomaly map and pass it to the classifier. The classifier only contains a convolution layer followed by a global average pooling layer and a fully-connected layer.

  \subsection{Loss Function}
  We first give the overall loss function for supervised training where the forgery masks are available, and then describe the weakly supervised training strategy using our single side loss.

  \textbf{Supervised Training.}
  let $\textbf{M} \in \{0,1\}^{H \times W}$ denotes the forgery mask, we divide it into patches $\textbf{m}_{i,j}$ (16 $\times$ 16 pixels in our case), we use its averaged value $AVG(\textbf{m}_{i,j})$ to represent the forgery score of $(i^{th},j^{th})$ patch in the face image. We use a \emph{hard thresholding} fashion to calculate the first and second order anomaly scores of two adjacent patches as shown in \eqref{eq:first} and \eqref{eq:second} respectively. Without loss of generality, we use $(i^{\prime},j^{\prime})$ to denote a neighbor of $(i,j)$.
  \begin{subequations}
    \small
    \begin{align}
      \label{eq:first}
      &a_{i,j}=
      \begin{cases}
            1& \text{if} \phantom{se} |AVG(\textbf{m}_{i,j}) - AVG(\textbf{m}_{i',j'})|>0, \\
            0& \text{else}.  
      \end{cases}\\ 
      \label{eq:second}
      &a^{\prime}_{i,j}=
      \begin{cases}
              1& \text{if} \phantom{se} |a_{i,j} - a_{i',j'}|>0, \\
              0& \text{else}. 
      \end{cases}
    \end{align}
  \end{subequations}
  By iterating \eqref{eq:first} and \eqref{eq:second} for each $\textbf{m}_{i,j}$, we obtain the ground truth of the first order anomaly maps $\tilde{\mathcal{M}} \in \{0, 1\}^{h \times w  \times 64}$ (repeated 64 times), and the second order anomaly maps $\tilde{\mathcal{M}}^{\prime} \in \{0, 1\}^{h^{\prime} \times w^{\prime}} $. The local anomaly maps can be optimized using pixel wise binary cross entropy (BCE):
  \begin{equation}
    \small
    \setlength{\abovedisplayskip}{5pt}
    \mathcal{L}_{A} =  \beta \sum_{*} BCE(\mathcal{M}_{*}, \tilde{\mathcal{M}}_{*}) + \gamma \sum_{h} BCE(\mathcal{M}^{\prime}_{*}, \tilde{\mathcal{M}}^{\prime}_{*}),
    \label{eq:sup}
    \setlength{\belowdisplayskip}{3pt}
  \end{equation}
  where $* \in \{v1,v2,h1,h2\}$, $\beta$ and $\gamma$ are the weights for the first and second order anomaly maps. The overall loss for training our model is:
  \begin{equation}
    \setlength{\abovedisplayskip}{3pt}
    \mathcal{L}_{total} = \alpha L_{cls} + L_{A},
    \setlength{\belowdisplayskip}{3pt}
    \label{eq:loss-unsuper}
  \end{equation}
  $L_{cls}$ is the BCE loss for the classifier and $\alpha$ is the weight.

  \textbf{Weakly Supervised Training.}
  Although SOLA learning module predicts patch level anomalies, we can train it in a weakly supervised manner where only image level labels are available. To this end, we have two hypotheses: (1) The pixels of anomaly maps of real faces should be all zero or close to zero. In the meantime, there should be a portion of nonzero pixels in the anomaly maps of forged faces. (2) Anomaly maps vary according to different forgery algorithms, but they should be always nonzero. On these hypotheses, we introduce a single side loss in Eq. \ref{eq:foa} that only penalizes the anomaly maps of real faces. This loss ensures multiple patterns of anomalies in forged faces, thereby improving the generalization:
  \begin{equation}
    \small
    \setlength{\abovedisplayskip}{4pt}
    \mathcal{L}_{A} = \sum_{I \in real} \sum_{*} \underbrace{\beta  ||\mathcal{M}_{*} -  \tilde{\mathcal{M}}_{*}||_1}_{\text{First Order}} + \underbrace{\gamma  ||\mathcal{M}^{\prime}_{*} - \tilde{\mathcal{M}}^{\prime}_{*}||_1}_{\text{Second Order}} ,
    \label{eq:foa}
    \setlength{\belowdisplayskip}{4pt}
  \end{equation}
  where $\tilde{\mathcal{M}}$ and $\tilde{\mathcal{M}}^{\prime}$ are both all-zero.

\section{Experiments}
\label{sec:Exp}

  \subsection{Experiments details}
  \textbf{Datasets.} We use a wide range of popular face forgery datasets to evaluate the proposed method, including FaceForensics++ (FF++) \cite{rossler2019faceforensics++}, Celeb-DF v2 (CD2) \cite{li2020celeb}, DeepfakeDetection Dataset (DFD) \cite{dfd}, and FaceShifter (Fshi) \cite{li2020advancing}. In FF++, 1000 original videos are forged by four forgery algorithms: DeepFakes, Face2Face \cite {thies2016face2face}, FaceSwap, and NeuralTextures \cite{thies2019deferred}. All face patches are cropped according to their masks if available, otherwise, we use RetinaFace \cite{deng2020retinaface} to detect faces and crop the patches. To preserve enough background, the cropped patches are set to be 2.6 times the size of masks or bounding box. All face patches are resized to 256 $\times$ 256 and normalize to $[0,1]$ by dividing 255.  

  \textbf{Evaluation Metrics.} 
  We report frame level AUC (area under the receiver operating characteristic curve) as most of the previous works do. 
  The experimental results of other methods which we use for comparison are directly cited.

  \textbf{Implementation Details.} 
  All the experiments are implemented with Pytorch \cite{paszke2019pytorch} with 4$\times$NVIDIA RTX 3090 24GB. The backbone is initialized with ImageNet pre-trained weights and trained using Adam optimizer \cite{kingma2014adam} with learning rate 1e-3, betas 0.9 and 0.999, and epsilon 1e-8. The batch size is 32 and number of epochs is 50 without early stopping.


  

  \begin{table}[htbp]
    \renewcommand{\arraystretch}{1.1}
    \small
    \centering
    \begin{tabular}{>{\centering}p{70pt}>{\centering}p{28pt}>{\centering}p{17pt}>{\centering}p{17pt}>{\centering}p{17pt}>{\centering\arraybackslash}p{17pt}}
    \toprule
    Methods                         &Backbone & DF              & F2F               &FS              &NT     \\ \hline
    Xception \cite{rossler2019faceforensics++}                  &Xception &99.38           &99.53               &99.36           &99.50 \\
    PBD \cite{schwarcz2021finding}                  &Xception &97.00           &95.00               &98.00          &98.00 \\ 
    Face X-ray \cite{li2020face}  &HRNet    &99.17              &99.06            &99.20           &98.93 \\
    S-MIL \cite{li2020sharp}      &Xception &99.84              &99.34            &99.61           &98.85\\\hline
    SOLA \emph{-weakly sup}     &ResNet18 &\textbf{100}       &\textbf{99.67}   &\textbf{100}    &\textbf{99.82}\\ 
    SOLA \emph{-sup}     &ResNet18 &100       &99.56   &99.98    &99.76\\ 
    \bottomrule
    \end{tabular}
    \caption{In-domain performences on FF++.}
    \label{table:ff-in-datset-raw}
  \end{table}

  \definecolor{mygray}{gray}{.9}
  \begin{table*}
    \small
    \centering
    \begin{tabular}{>{\centering}p{85pt}>{\centering}p{50pt}>{\centering}p{40pt}>{\centering}p{35pt}>{\centering}p{35pt}>{\centering}p{35pt}>{\centering}p{35pt}>{\centering\arraybackslash}p{35pt}}
    \toprule
    \multirow{2}{*}{Model} & \multirow{2}{*}{Backbone} & \multirow{2}{*}{Train Set} & \multicolumn{4}{c}{Test Set} & \multirow{2}{*}{Avg}  \\\cline{4-7} 
                           &                           &                            & DF    & F2F  & FS    & NT    &      \\ \hline\hline
    Xception \cite{rossler2019faceforensics++} & Xception     & \multirow{4}{*}{DF} &\multicolumn{1}{>{\columncolor{mygray}}c}{99.38} &75.05  &49.13  &80.39   &75.99 \\
    Face X-ray \cite{li2020face}               & HRNet        &                     &\multicolumn{1}{>{\columncolor{mygray}}c}{99.17} &94.14   &\textbf{75.34}  &93.85  &90.63 \\
    SOLA  \emph{-weakly sup}                & ResNet18              &                     &\multicolumn{1}{>{\columncolor{mygray}}c}{100}     &\textbf{97.29}   &63.59  &98.45  &89.83 \\ 
    SOLA \emph{-sup}                  & ResNet18                &                     &\multicolumn{1}{>{\columncolor{mygray}}c}{\textbf{100}}      &96.95   &69.72   &\textbf{98.48} &\textbf{91.28} \\\hline\hline
    
    Xception \cite{rossler2019faceforensics++}           & Xception     & \multirow{4}{*}{F2F}        &87.56 &\multicolumn{1}{>{\columncolor{mygray}}c}{99.53}   &65.23 &65.90  &79.56 \\
    Face X-ray  \cite{li2020face}          & HRNet     &       &98.52 &\multicolumn{1}{>{\columncolor{mygray}}c}{99.06}   &72.69   &91.49 &90.44       \\
    SOLA   \emph{-weakly sup}             & ResNet18                 &                            &99.61 &\multicolumn{1}{>{\columncolor{mygray}}c}{\textbf{99.67}}   &84.24 &\textbf{97.48}  &95.25 \\
    SOLA \emph{-sup}                   & ResNet18                 &                             &\textbf{99.73} &\multicolumn{1}{>{\columncolor{mygray}}c}{99.56}   &\textbf{93.50} &96.02  &\textbf{97.20} \\ \hline\hline

    Xception \cite{rossler2019faceforensics++}           & Xception     & \multirow{4}{*}{FS}        &70.12 &61.70   &\multicolumn{1}{>{\columncolor{mygray}}c}{99.36} &68.71  &74.97 \\
    Face X-ray  \cite{li2020face}          & HRNet     &       &93.77 & 92.29   &\multicolumn{1}{>{\columncolor{mygray}}c}{99.20}   &86.63 &92.97      \\
    SOLA  \emph{-weakly sup}             & ResNet18                 &                            &93.18 &97.59   &\multicolumn{1}{>{\columncolor{mygray}}c}{\textbf{100}} &94.93  &96.43 \\
    SOLA \emph{-sup}                  & ResNet18                 &                             &\textbf{99.11} &\textbf{98.13}   &\multicolumn{1}{>{\columncolor{mygray}}c}{99.98} &\textbf{92.07}  &\textbf{97.32} \\ \hline\hline

    Xception \cite{rossler2019faceforensics++}         & Xception     & \multirow{4}{*}{NT}        &93.09 &84.82   &47.98  &\multicolumn{1}{>{\columncolor{mygray}}c}{99.50}  &81.35 \\
    Face X-ray \cite{li2020face}                        & HRNet       &                            &99.14 &\textbf{98.43}   &70.56  &\multicolumn{1}{>{\columncolor{mygray}}c}{98.93}  &91.77       \\
    SOLA  \emph{-weakly sup}             & ResNet18                      &                            &\textbf{99.95} &94.83   &57.32  &\multicolumn{1}{>{\columncolor{mygray}}c}{\textbf{99.82}} &87.97  \\ 
    SOLA \emph{-sup}                    & ResNet18                  &                            &99.64 &97.69   &\textbf{90.20}  &\multicolumn{1}{>{\columncolor{mygray}}c}{99.76}  &\textbf{96.82}  \\ 
    \bottomrule
    \end{tabular}
    \caption{Cross-domain evaluations on FF++. }
    \label{table:ff-cross-datset-raw}
  \end{table*}

  \subsection{In-Domain Evaluations}

  We first report the performance of our method under in-domain evaluations. The results of FF++ are shown in Table \ref{table:ff-in-datset-raw}. 
  \emph{-weakly sup} and \emph{-sup} denote SOLA trained with different strategies.
  Previous methods have achieved great performance, but our method still surpasses the best competitors by about 1\% in terms of average AUC .
  Note that our method only uses a relatively small backbone ResNet18 while most of other methods use a large backbone like Xception. 
  On the whole, weakly supervised SOLA performs slightly better, proving that the single side loss allows SOLA to capture different anomalies precisely.

  \subsection{Cross-domain Evaluations}
  In this section, we focus on the more challenging cross-domain evaluations to explore the performances of our method on unseen forgeries. 
  Table \ref{table:ff-cross-datset-raw} shows the results of cross-domain evaluations on FF++. 
  Here, the models are trained on only one dataset and evaluated on all four datasets. 
  Although Xception achieves 99.42\% average AUC of in-domain evaluations, it can not generalize to other forgeries well and its performances of cross-domain evaluations drop sharply. 
  Meanwhile, Face x-ray \cite{li2020face} is designed to detect the blending boundaries instead of specifical artifacts caused by forgery algorithms and made significant progress in generalization. 
  We can see that weakly supervised SOLA performs closely to Face x-ray.  
  Face x-ray requires pixel level forgery masks to locate the boundaries and train their model, but our method also achieves a great improvement of generalization ability with only image level labels. 
  With the help of pixel level annotations, supervised SOLA surpasses all other methods in terms of average AUC of cross-domain evaluations. The overall performance of supervised SOLA on FF++ exceeds Xception and Face x-ray by 17.69\% and 4.21\%.

  \begin{table}[htbp]
    \centering
    \renewcommand{\arraystretch}{1.1}
    \begin{tabular}{>{\centering}p{85pt}>{\centering}p{35pt}>{\centering}p{35pt}>{\centering\arraybackslash}p{35pt}}
    \toprule
    \multirow{2}{*}{Model}                  & \multirow{2}{*}{Backbone}                 &\multirow{2}{*}{Train Set}            & Test Set    \\ \cline{4-4} 
                                                &                  &             &FShi    \\ \hline
    Xception    \cite{rossler2019faceforensics++}            &     Xception             &     FF++             &72.00         \\ 
    PBD        \cite{schwarcz2021finding}           &     Xception             &        FF++           &  57.80       \\ 
    FWA  \cite{li2019exposing}          &    ResNet152              &         FF++          &   65.50      \\
    Face X-ray  \cite{li2020face}          &    HRNet              &         FF++          &   92.80      \\
    LipForensics \cite{haliassos2021lips}            &    ResNet18              &         FF++          &   97.10      \\\hline
    SOLA \emph{-weakly sup} &   ResNet18       &   FF++                 &97.27       \\ 

    SOLA \emph{-sup}  &  ResNet18           &  FF++                  &\textbf{98.72}       \\      
    \bottomrule
    \end{tabular}
    \caption{Cross-domain evaluation results on FaceShifter (trained on FF++). Our method has better performance than state-of-the-art works and is also competitive while using only one dataset.}
    \label{table:cross-datset-fshi}
  \end{table}

  \begin{table}[htbp]
    \renewcommand{\arraystretch}{1.1}
    \centering
    \begin{tabular}{>{\centering}p{95pt}>{\centering}p{60pt}>{\centering\arraybackslash}p{45pt}}
    \toprule
    \multirow{2}{*}{Model}                  &  \multirow{2}{*}{Train Set}                    & Test Set    \\ \cline{3-3} 
                                            &                & CD2    \\ \hline
    $F^{3}$ Net \cite{qian2020thinking}   &      FF++(c23)                 & 65.17        \\
    FWA  \cite{li2019exposing}         &     Self-made                        & 57.32        \\ 
    MADD    \cite{zhao2021multi}               &    FF++(c23)                         & 67.44         \\
    MTD-Net  \cite{yang2021mtd}        &      FF++(c23)                 & 70.12        \\ 
    Two Branch \cite{masi2020two}     &      FF++(c40)                 & 73.41        \\  
    F3Net \cite{li2021frequency}     &      FF++(c23)                 & 65.20        \\ 
    LRL  \cite{chen2021local}     &      FF++(c23)                 & \textbf{78.26}        \\ 
    GFF  \cite{luo2021generalizing}     &      FF++(c23)                 & 65.20        \\ 

    MADD  \cite{zhao2021multi}     &      FF++(c23)                 & 67.44        \\
    SPSL \cite{liu2021spatial}     &      FF++(c23)                 & 76.88        \\
    LipFor   \cite{haliassos2021lips}     &      FF++(c23)                 & 82.40        \\
    \hline
    SOLA  \emph{-weakly sup}              & DF(c23)                              &72.47         \\  
    SOLA  \emph{-sup}  & DF(c23)                              &76.02           \\  
    \bottomrule
    \end{tabular}
    \caption{Cross-domain evaluations on CD2. Our methods trained using only DF-c23 outperforms most methods.}
    \label{table:cross-datset-cd}
  \end{table}

  Results of cross-domain evaluations on a more advanced face forgery algorithm FaceShifter (Fshi) are shown in Table \ref{table:cross-datset-fshi}. 
  Following other works, we train our model on FF++ and evaluate it by Fshi. 
  Weakly supervised and supervised SOLA both exceed state-of-the-art method, by 0.17\% and 1.62\% AUC respectively.
  In addition to the evaluations with two supervision fashions, we also evaluate weakly supervised SOLA trained with only one dataset. 
  Although the performances of weakly supervised SOLA are slightly lower than the supervised SOLA when trained using only one dataset, they are generally still better than most of the recent methods.
  These results demonstrate the local anomaly traces are universal in the forged faces and can be generalized to unseen forgeries.

  The comparisons of cross-domain evaluation on CD2 are given in Table \ref{table:cross-datset-cd}. Note that the compared methods are trained with different datasets, so we present the results for reference only. Most methods use all four datasets of FF++ so as to obtain good generalization on CD2. 
  However, weakly supervised SOLA gets 72.47\% AUC when trained using only the compressed DF dataset, exceeding most state-of-the-art competitors. Supervised SOLA further improve the result to 76.02\% which is 3.55\% higher than the Two-Branch \cite{masi2020two} that similarly fuses RGB and frequency features. 
  \begin{table}[htbp]
    \centering
    \small
    \renewcommand{\arraystretch}{1.1}
    \begin{tabular}{>{\centering}p{70pt}>{\centering}p{28pt}>{\centering}p{17pt}>{\centering}p{17pt}>{\centering}p{17pt}>{\centering\arraybackslash}p{17pt}}
    \toprule
    \multirow{2}{*}{Model} & \multirow{2}{*}{Train Set} & \multicolumn{4}{c}{Test Set} \\ \cline{3-6} 
                           &                      & DF    & F2F   & FS   & NT        \\ \hline
    Xception\cite{rossler2019faceforensics++}                  &   \multirow{2}{*}{CD2}                 &\textbf{87.69}   &75.17  &54.19 &72.89    \\ 
    SOLA    \emph{-weakly sup}               &                    &85.77   &\textbf{85.72}  &\textbf{86.78} &\textbf{85.64}    \\ \hline
    Xception\cite{rossler2019faceforensics++}                &   \multirow{2}{*}{DFD}                &94.79   &76.96  &47.13 &84.91    \\ 
    SOLA   \emph{-weakly sup}                &                   &\textbf{95.51}   &\textbf{85.51}  &\textbf{60.81} &\textbf{84.01}    \\ 
    \bottomrule
    \end{tabular}
    \caption{Cross-domain evaluations on DF, F2F, FS, and NT by training on CD2 and DFD.}
    \label{table:cross-datset-cd-dfd}
  \end{table}

  Table \ref{table:cross-datset-cd-dfd} shows the cross-domain results of our method trained on CD2 and DFD and tested on FF++. 
  Here, SOLA can be only trained without supervision given that the ground truth forgery masks of both the datasets are not available. 
  The overall AUCs are 85.98\% and 81.46\% when trained on CD2 and DFD respectively, exceeding the Xception by 13.49\%, 5.57\%, proving that our method generalizes well on multiple datasets even with only image level annotations.

  \begin{figure}[htbp]
    \centering
     \includegraphics[width=\linewidth]{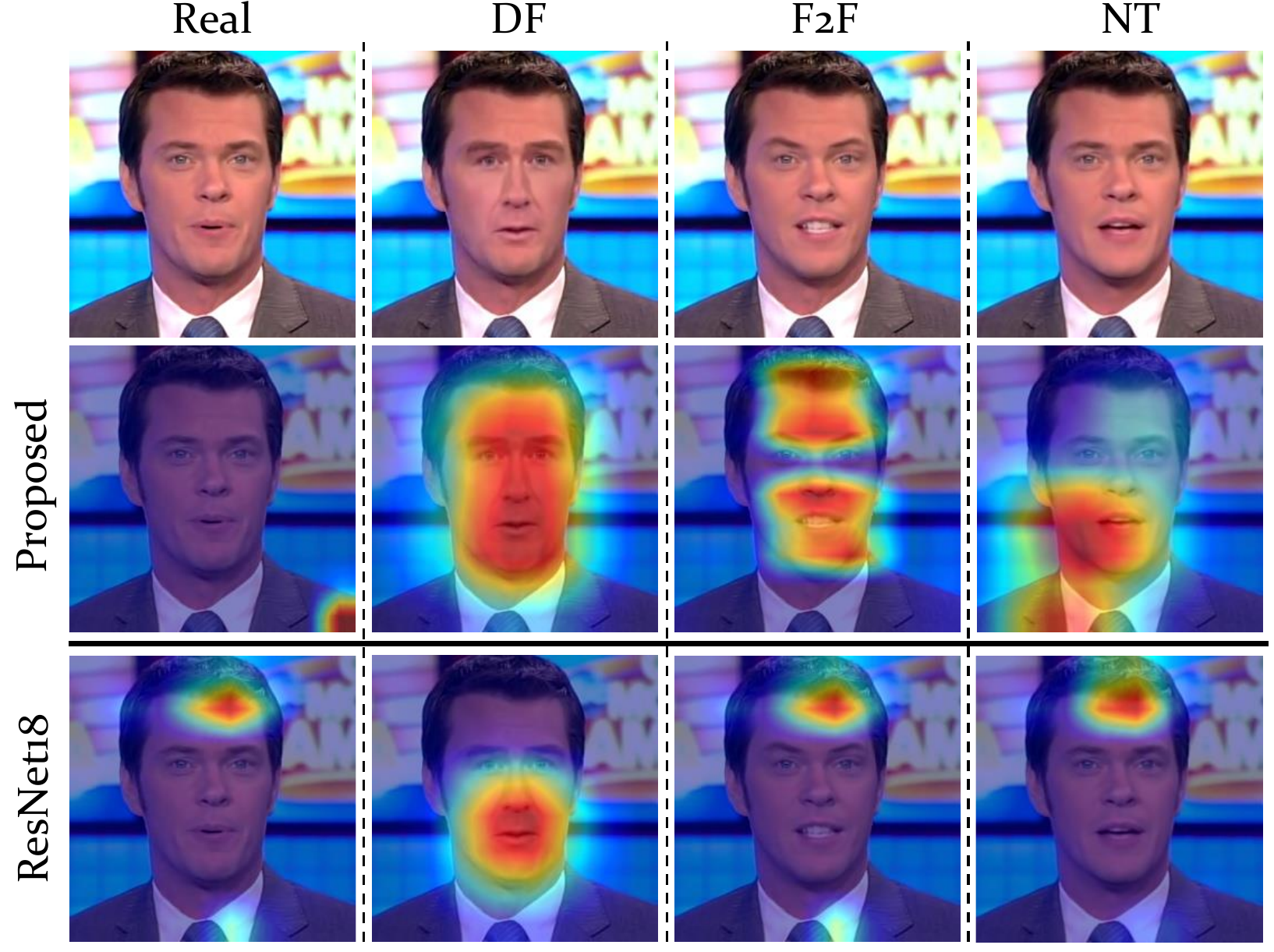}
     \caption{Grad-CAM \cite{selvaraju2017grad} for the \emph{forgery} class of different of forged faces. We can see the proposed SOLA can respond to various forgeries while the vanilla ResNet18 can not.}
     \label{fig:cam}
  \end{figure}



  To explore which region our method pays attention when encountering unseen forgeries, we use Grad-CAM \cite{selvaraju2017grad} to generate heat maps as illustrated in fig. \ref{fig:cam}. 
  The warm color marks the regions that respond strongly to the prediction of \emph{forgery}. 
  Here, a ResNet18 and our method are both trained on DF.  
  Obviously, they can well concentrate on the forged regions of faces created by DF and have almost no response to real faces. 
  However, our method not only captures forged regions on DF more comprehensively, but also responds to unseen forgeries while ResNet18 can not. 
  We can see that ResNet18 fails to capture the forged regions on F2F and NT, thus it responses the same to F2F and NT as to the real face. This agrees with the weak performance of deep learning models in cross-domain evaluations and reveals the generalization ability of our method on unseen forgeries.

  \begin{figure*}[thp]
    \centering
     \includegraphics[width=\linewidth]{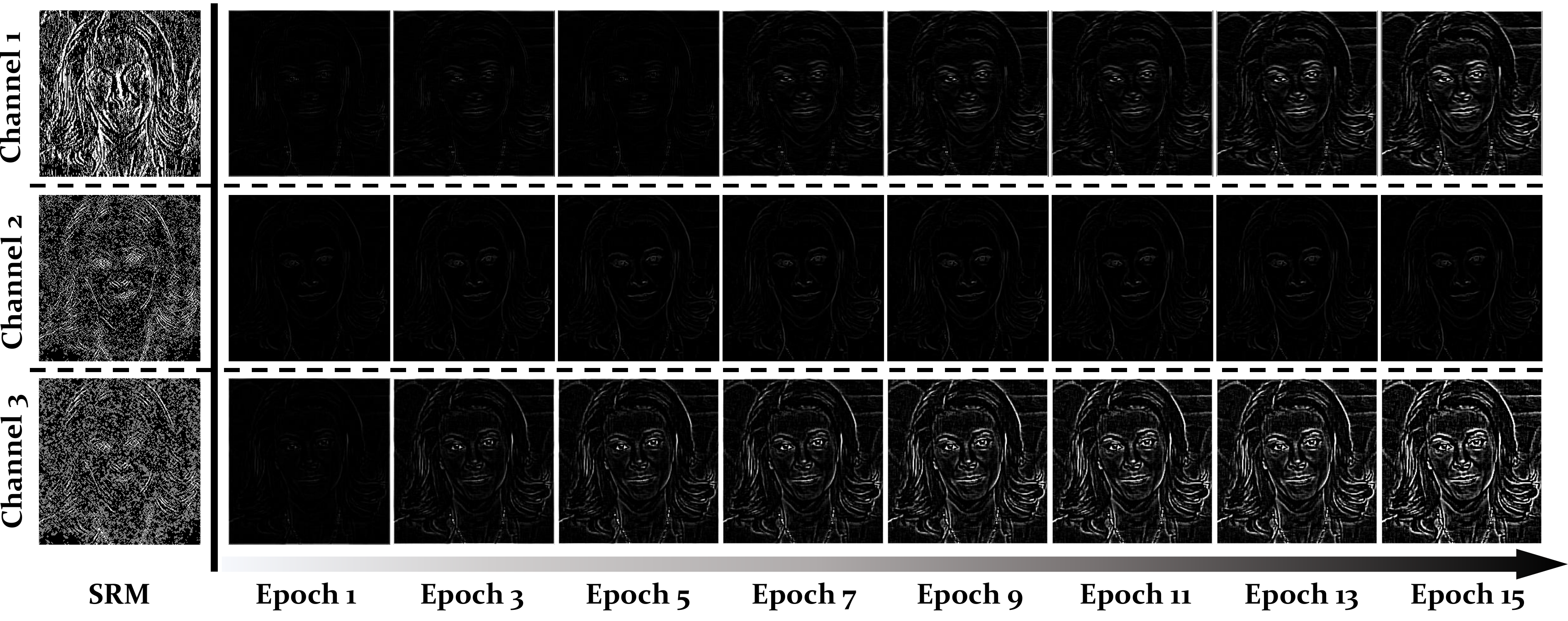}
     \caption{The outputs of different SRM variants over training epochs.}
     \label{fig:noise}
  \end{figure*}

  \subsection{Ablation Studies}
  \textbf{The effect of ASRM.}
  So far, we have been training our method with the proposed ASRM. In this section, we demonstrate its effectiveness by replacing it with several variants. 
  Table \ref{fig:noise} shows the performances of different noise extraction strategies in the noise branch: \emph{-w/o noise} denotes SOLA without noise branch, \emph{-srm} and \emph{-lsrm} denote SOLA using fixed SRM filters and learnable SRM filters (use filters as initialization of the first layer without any constraint). 
  All variants are trained on CD2 and DFD without pixel level annotations. 
  The results show that the noise branch effectively improves the performance of SOLA on DFD, by 16.5\%, 16.68\%, and 16.67\% with different SRM variants.
    \begin{table}[htbp]
    \centering
    \small
    \renewcommand{\arraystretch}{1.1}
    \begin{tabular}{>{\centering}p{70pt}>{\centering}p{27pt}>{\centering\arraybackslash}p{27pt}}
    \toprule
    %
    Model                     & CD2             & DFD     \\ \hline
    SOLA \emph{-w/o noise}     &98.15              &83.31    \\ 
    SOLA \emph{-srm}       &73.05              &99.81 \\ 
    SOLA \emph{-lsrm}      &94.67              &\textbf{99.99}  \\ 
    SOLA \emph{-asrm}      &\textbf{98.70}     &99.98   \\ 
    \bottomrule
    \end{tabular}
    \caption{Comparsion results of different variants of SRM.}
    \label{table:cross-datset-auc-ap-eer}
  \end{table}

  Fig. \ref{fig:noise} shows outputs of all three channels of standard SRM, LSRM, and ASRM across training epochs. The input face is the same as that in Fig. \ref{fig:srm}. 
  Although the standard SRM extracts high frequency noise, it does not focus on forged regions. 
  Meanwhile, the LSRM loses its high pass peculiarity and fails to extract noise features in the training process. 
  But we can see that ASRM gradually extracts the high frequency noise of face regions, especially the facial organs, providing discriminative cues that can not be learned from CNNs directly.

  \textbf{The effect of LEM.}
  To confirm the effect of LEM, we train the weakly supervised SOLA with different combinations of ASRM and LEM. As shown in Table \ref{table:cross-datset-cd-size}, the models are trained on DF and evaluated on DF, FShi, and DFD. Generally, both ASRM and LEM can improve the overall performances while ASRM contributes more to the generalization and increases the AUC on Fshi and DFD by 8.78\% and 10.16\%.

  \textbf{The effect of Patch Size.}
  In the previous experiments, we use the 3 out of 4 blocks in ResNet18 and an additional pooling layer to obtain feaures of size $16 \times 16 \times 256$. To further evaluate the effect of different patch sizes, we change the pooling size to obtain features of size $8 \times 8$ and  $32 \times 32$. 
  The results are shown in Table \ref{table:cross-datset-cd-size}.
  Although SOLA with different patch sizes all get good results, SOLA with patch size 16 achieves the optimal performances across all datasets. These results accord with the conclusions in \cite{zhao2021learning} as well.

    \begin{table}[htbp]
    \renewcommand{\arraystretch}{1.1}
    \centering
    \begin{tabular}{>{\centering}p{28pt}>{\centering}p{28pt}>{\centering}p{40pt}>{\centering}p{22pt}>{\centering}p{22pt}>{\centering\arraybackslash}p{22pt}}
    \toprule
    \multirow{2}{*}{ASRM} &\multirow{2}{*}{LEM} & \multirow{2}{*}{Patch Size} & \multicolumn{3}{c}{Test Set} \\ \cline{4-6} 
                                  &                    &                                &    DF     &FShi  &DFD       \\     \hline
       &                    &       16                       &  99.90   & 72.24 &   78.33   \\     
        $\surd$                   &                    &       16                       &  99.87   & 81.02 &   88.49   \\     
                                  &        $\surd$            &    16                   &  99.98   & 73.64 & 80.25     \\     \hline
    $\surd$                         &  $\surd$                &  8                      &100       &80.17  &91.10    \\ 
    $\surd$                       &   $\surd$              &   16                       &\textbf{100}    &\textbf{86.53}       &\textbf{92.61}     \\
    $\surd$                         &  $\surd$                &  32                     &100        &85.99   &87.85  \\      
    \bottomrule
    \end{tabular}
    \caption{Performances of weakly supervised SOLA with different patch size and the effect of LEM.}
    \label{table:cross-datset-cd-size}
  \end{table}

  \begin{table}[htbp]
    \centering
    \small
    \renewcommand{\arraystretch}{1.1}
    \begin{tabular}{>{\centering}p{65pt}>{\centering}p{35pt}>{\centering}p{35pt}>{\centering\arraybackslash}p{35pt}}
    \toprule
    %
    Backbone                     & FF++(c0)             & FF++(c23)     & CD2   \\ \hline
    ResNet18                     &99.84              &98.10           & 68.33   \\ 
    ResNet50                    &99.90              &99.14            & 74.98  \\ 
    ResNet101                    &\textbf{99.90}            &\textbf{99.25}  & \textbf{75.05}  \\ 
    \bottomrule
    \end{tabular}
    \caption{Comparsion results of different backbones.}
    \label{table:Backbones}
  \end{table}

  \textbf{The effect of Backbones.} We further test the effect of different backbone models on both in-domain and cross domain settings. As in Table \ref{table:Backbones}, we choose 2 other models in ResNet family and report the results on FF++ and CD2 (trained on FF++(c23)). While the overall performances especially the cross-domain performance are much better with ResNet50, the increase is very limited with a deeper ResNet101.

  \subsection{Limitations}
  While the proposed method predicts fine-grained local anomalies of different distances and directions, the diversity of different predictors in SOLA is not explicitly ensured, especially under the weakly-sup setting. This could lead to the degeneration of local anomaly predictors and weaken the representations of local anomalies.

\section{Conclusion}
\label{sec:conclusion}
In this work, we revisit face forgery detection from the perspective of local anomaly detection and propose the SOLA learning module to predict different types of local anomalies in the first and second order. Besides, we design a weakly supervised strategy to train SOLA without pixel level annotations. We also introduce a adaptive spatial rich model to mine subtle high frequency traces using learnable high pass kernels. Experiments on multiple datasets demonstrate that our method achieves competitive performance with a small backbone and generalize to unseen forgery types well.

\section*{Acknowlegdement}
\label{sec:acknowlegdement}
This work is supported in part by the National Key Research and Development Plan of China under Grant 2020YFB1005600, in part by the National Natural Science Foundation of China under grant numbers 62122032, and U1936118, in part by Qinglan Project of Jiangsu Province, and ``333'' project of Jiangsu Province.

{
  \small
  \bibliographystyle{ieee_fullname}
  \bibliography{egbib}
}

\end{document}